\documentclass[10pt,twocolumn,letterpaper]{article}

\usepackage{cvpr}              

\makeatletter
\DeclareRobustCommand\onedot{\futurelet\@let@token\@onedot}
\def\@onedot{\ifx\@let@token.\else.\null\fi\xspace}

\def\ie{\emph{i.e}\onedot} 
\def\cf{\emph{cf}\onedot}

\makeatother

\usepackage{multirow}
\usepackage{color}
\usepackage{colortbl}
\definecolor{cvprblue}{rgb}{0.21,0.49,0.74}
\usepackage[pagebackref,breaklinks,colorlinks,allcolors=cvprblue]{hyperref}
\usepackage[accsupp]{axessibility} 

\def\paperID{39873} 
\def\confName{CVPR}
\def\confYear{2026}

\title{RU4D-SLAM: Reweighting Uncertainty in Gaussian Splatting SLAM \\ for 4D Scene Reconstruction}

\author{\textbf{Yangfan Zhao}$^1$\thanks{Equal contribution}  ~~~ \textbf{Hanwei Zhang}$^2$\footnotemark[1] ~~~ \textbf{Ke Huang}$^3$ \\ 
\textbf{Qiufeng Wang}$^3$ ~~~ \textbf{Zhenzhou Shao}$^{1\dag}$ ~~~ \textbf{Dengyu Wu}$^{4\dag}$\\
$^1$Capital Normal University, $^2$Saarland University\\
$^3$Xi'an Jiaotong-Liverpool University, $^4$King's College London\\
$^*$Equal contribution ~~~ $\dag$ Corresponding Author \\
}

\begin{document}


\twocolumn[{%
\renewcommand\twocolumn[1][]{#1}%
\maketitle
\vspace{-3em}
\begin{center}
    \includegraphics[width=0.98\textwidth]{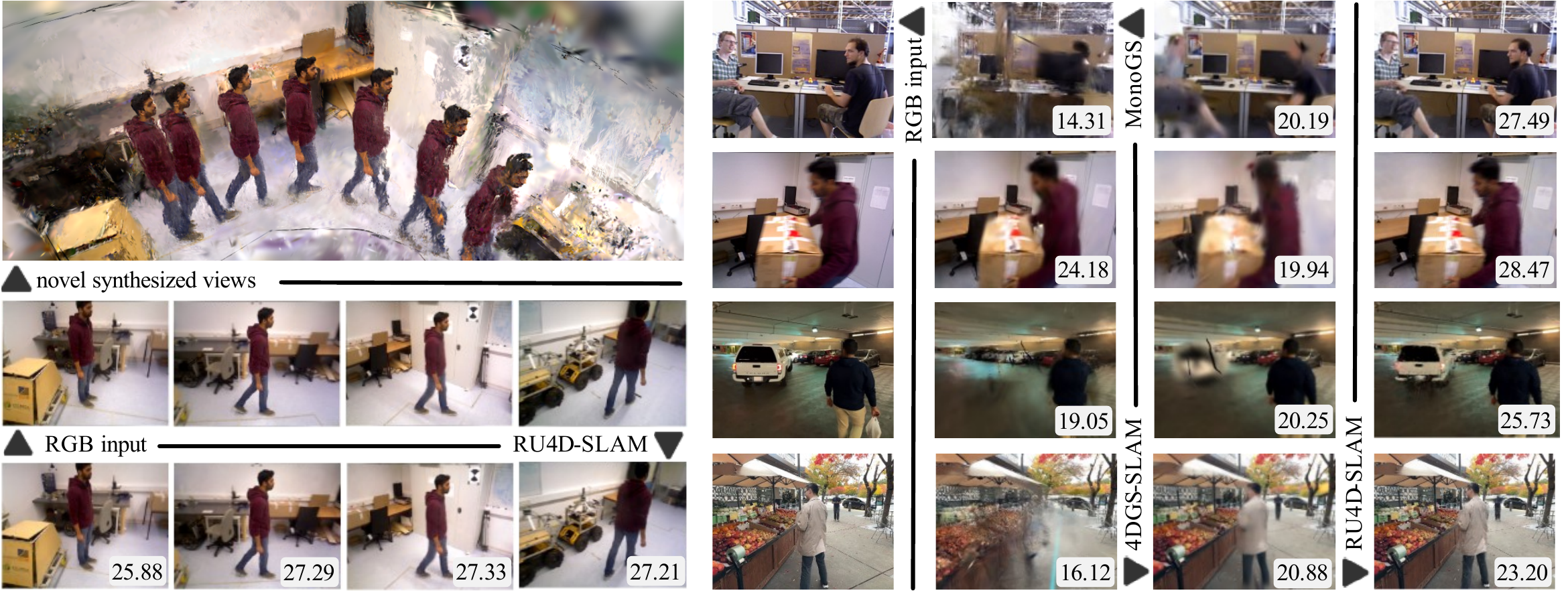}
    \vspace{-1.em}
    \captionof{figure}{\textbf{4D scene reconstruction with RU4D-SLAM.} 
    The left side shows the 4D Gaussian map reconstructed by RU4D-SLAM on the Bonn dataset~\cite{bonn_rgbd_dynamic_dataset}, featuring novel synthesized views that capture the temporal motions (top) and a comparison between the RGB input and RU4D-SLAM rendering at the same pose (bottom).
    On the right, we compare rendered results from MonoGS~\cite{matsuki2024gaussian}, 4DGS-SLAM~\cite{li20254d}, and RU4D-SLAM in dynamic scenes. The numbers at the bottom-right of each image denote the PSNR values (higher is better).}
    \label{fig:teaser}
\end{center}
}]

\begin{abstract}
Combining 3D Gaussian splatting with Simultaneous Localization and Mapping (SLAM) has gained popularity as it enables continuous 3D environment reconstruction during motion. However, existing methods struggle in dynamic environments, particularly moving objects complicate 3D reconstruction and, in turn, hinder reliable tracking.
The emergence of 4D reconstruction, especially 4D Gaussian splatting, offers a promising direction for addressing these challenges, yet its potential for 4D-aware SLAM remains largely underexplored.
Along this direction, we propose a robust and efficient framework, namely Reweighting Uncertainty in Gaussian Splatting SLAM (RU4D-SLAM) for 4D scene reconstruction, that introduces temporal factors into spatial 3D representation while incorporating uncertainty-aware perception of scene changes, blurred image synthesis, and dynamic scene reconstruction. We enhance dynamic scene representation by integrating motion blur rendering, and improve uncertainty-aware tracking by extending per-pixel uncertainty modeling, which is originally designed for static scenarios, to handle blurred images. Furthermore, we propose a semantic-guided reweighting mechanism for per-pixel uncertainty estimation in dynamic scenes, and introduce a learnable opacity weight to support adaptive 4D mapping.
Extensive experiments on standard benchmarks demonstrate that our method substantially outperforms state-of-the-art approaches in both trajectory accuracy and 4D scene reconstruction, particularly in dynamic environments with moving objects and low-quality inputs. Code available: \href{https://ru4d-slam.github.io}{https://ru4d-slam.github.io}
\vspace{-1.2em}
\end{abstract}

\section{Introduction}
Simultaneous Localization and Mapping (SLAM)~\cite{mur2015orb,matsuki2024gaussian} along with 3D reconstruction~\cite{10.1145/3592433}, or more recent 4D reconstruction~\cite{wang2023badnerf,zhou2024drivinggaussian}, which integrates trajectory estimation with the modeling of spatio-temporal structures, have emerged as fundamental approaches for environmental representation. 
SLAM with 3D or 4D reconstruction is essential to embodied intelligence, as accurate environmental perception and modeling enable agents to localize, navigate, and interact effectively with the world~\cite{long2025survey}.
Despite decades of progress, these tasks remain inherently difficult due to the complexity and variability of real-world environments. Challenges arise from dynamic objects that disrupt static scene reconstruction, as well as visual degradation caused by motion blur and inconsistent exposure, including both overexposure and underexposure.
These factors introduce spatio-temporal uncertainty, posing significant difficulties for current perception pipelines.

Classical SLAM pipelines, such as ORB-SLAM~\cite{mur2015orb}, demonstrate accurate localization in static environments, while dense reconstruction systems like KinectFusion~\cite{newcombe2011kinectfusion} enable detailed 3D mapping. 
Building on these foundations, neural implicit methods~\cite{sucar2021imap,Zhu_2022_CVPR} and Gaussian splatting approaches~\cite{yan2024gs,matsuki2024gaussian} have further advanced scene fidelity and scalability by capturing richer information. 
As a result, 3D reconstruction quality in static settings has significantly improved.
However, extending these techniques to dynamic environments remains a major challenge. To address it, recent extensions such as DG-SLAM~\cite{xu2024dgslam} and WildGS-SLAM~\cite{Zheng_2025_CVPR} further improve tracking stability by masking or removing temporally inconsistent regions. Yet, these methods can still compromise static reconstruction, and WildGS-SLAM is unable to distinguish between dynamic objects and low-quality input data.

In parallel, 4D reconstruction has offered a new perspective for addressing dynamic environments by explicitly modeling spatio-temporal variations.
Early dynamic NeRFs ~\cite{pumarola2021d, tretschk2021non, fang2022fast, yang2022banmo, jiang2022neuman, weng2022humannerf} capture non-rigid deformations over time, but are computationally intensive. In contrast, Gaussian-based approaches~\cite{wu20244d, huang2024sc, hu2025learnable, wang2025freetimegs, li2024spacetime, park2025splinegs, lei2025mosca} achieve real-time rendering but rely on external pose estimates. Integrating 4D reconstruction with SLAM, as in 4D Gaussian splatting SLAM~\cite{li20254d}, improves both pose tracking and rendering quality in dynamic scenes. However, this represents only an initial step: the term ‘dynamic’ here refers solely to moving objects and overlooks challenges posed by low-quality input data, leaving significant room for further improvements. 

%

To advance 4D SLAM in dynamic environments, we extend the problem definition to include not only moving objects but also low-quality inputs caused by motion blur and inconsistent exposure. We introduce RU4D-SLAM, a 4D Gaussian splatting SLAM framework that explicitly separates static and dynamic regions using our \emph{reweighted uncertainty mask} (RUM). To further improve 4D mapping, we introduce two additional components: \emph{integrate and render} (IR), which enhances static mapping for stable localization and geometry under degraded inputs, and \emph{adaptive opacity weighting} (AOW), which improves dynamic mapping for moving objects. 

Conceptually, under the proposed uncertainty-driven framework, scene dynamics are represented by deformation nodes initialized within RUM regions as local motion anchors, while static regions remain undeformed. Both static and dynamic Gaussians are optimized using IR to handle motion blur and exposure inconsistency. These nodes evolve over time to capture non-rigid motion, and their trajectories are propagated to nearby Gaussians through deformation blending that interpolates neighboring transformations. AOW further modulates each deformed Gaussian’s contribution through learned temporal weights, stabilizing deformation propagation and enabling temporally consistent 4D reconstruction.

As shown in Figure~\ref{fig:teaser}, the method consistently improves reconstruction quality. RU4D-SLAM is capable of generating the synthesized view that captures the temporal dimension with decent quality. Besides, its renderings at given poses outperform state-of-the-art approaches in PSNR while remaining visually closest to the RGB inputs, even in the presence of motion blur, inconsistent exposure, and dynamic objects across both indoor and outdoor scenarios.
The main contributions of RU4D-SLAM are:
\begin{itemize}
\item We introduce a unified exposure-aware rendering formulation that accumulates rendering along camera trajectories, enabling motion-blur modeling while providing reliable uncertainty estimation in dynamic scenes.
\item We propose a reweighted uncertainty mask that combines exposure-driven reliability and semantic cues to distinguish dynamic and static regions, providing robust guidance for dynamic reconstruction.
\item We design an adaptive 4D mapping module that learns time-varying opacity and deformation fields guided by uncertainty, maintaining geometric consistency and temporal coherence under complex motion.
\end{itemize}

\section{Related Work}

\paragraph{Static 3D SLAM.}
Conventional SLAM systems estimate pose relying on geometric features~\cite{klein2007parallel,mur2015orb, qin2018vins, shan2021lvi, bescos2018dynaslam} or dense photometric alignment~\cite{engel2014lsd, engel2017direct, newcombe2011dtam, steinbrucker2011real, teed2021droid}. However, the visual sparsity and geometric rigidity of traditional SLAM pipelines limit their ability to capture dense and semantically rich 3D structures, motivating neural SLAM frameworks that integrate 3D reconstruction and pose estimation within a unified representation. For instance, NeRF-based systems~\cite{sucar2021imap, Zhu_2022_CVPR} demonstrated online implicit scene reconstruction but at significant computational cost, while voxel- and point-based representations~\cite{yang2022vox,johari2023eslam,wang2023co,sandstrom2023point} improved scalability through explicit spatial modeling, yet often sacrifice photorealism and fine-grained surface detail.
Most recently, 3D Gaussian splatting introduced a fast, differentiable representation that has led to Gaussian-based SLAM systems, such as SplaTAM~\cite{keetha2024splatam}, which jointly optimizes poses and Gaussian primitives, MonoGS~\cite{matsuki2024gaussian} with analytical Jacobians and regularization, and~\cite{huang2024photo}, which integrates Gaussian maps into feature-based frameworks. 
Continued improvements, including coarse-to-fine optimization~\cite{yan2024gs}, uncertainty modeling~\cite{hu2024cg}, and dynamic scene handling~\cite{xu2024dgslam}, aim to enhance robustness.
While static 3D SLAM has laid the foundation for robust mapping and localization, its inability to cope with dynamic changes in the environment necessitates the exploration of dynamic 3D SLAM, where both motion understanding and reconstruction are jointly addressed.

\paragraph{Dynamic 3D SLAM.} 
Dynamic SLAM systems aim to maintain robustness and reconstruct a consistent map when the environment contains dynamic objects.
Traditional approaches such as~\cite{bescos2018dynaslam, bescos2021dynaslam, liu2022rgb} remove dynamic objects through prior semantic or geometric masking, ensuring stable localization but discarding temporal information that could contribute to scene understanding. 
Building upon this foundation, recent research has progressed toward neural dynamic 3D SLAM, aiming to reconstruct and localize within dynamic environments jointly. ~\cite{jiang2024rodyn, xu2024dgslam} also incorporate prior semantic information mask, while ~\cite{li2025ddn} utilizes prior semantics to constrain the segmentation range of dynamic object masks and performs feature point segmentation based on a mixture of Gaussian distributions. 
However, these methods still depend heavily on pre-trained object detection or semantic segmentation networks, making them reliant on prior knowledge of specific object classes and thus limiting their generation and adaptability in complex real-world scenarios.
Therefore, building on 3D Gaussian splatting,~\cite{xu2024dgslam, wu2025add} use motion masks derived from depth to filter non-static areas, stabilizing pose estimation, while~\cite{jiang2024rodyn, li2025dy3dgs} incorporate optical flow estimation to recognize dynamic regions.~\cite{zhang2024monst3r, wang2024dust3r} detect moving objects by thresholding the difference between the predicted optical flow and the reprojection flow.~\cite{ren2024nerf, Zheng_2025_CVPR, zheng2025up} trains Gaussian masks to estimate uncertainty during inference for more resilient mapping. 
Despite advances, dynamic 3D SLAM still lacks robustness to low-quality inputs and the ability to model temporal scene changes, motivating 4D reconstruction and SLAM to capture geometry and motion jointly over time.

\paragraph{4D reconstruction and SLAM.}
Beyond static mapping, 4D reconstruction explicitly models temporal dynamics in scenes.~\cite{jiang2022neuman, weng2022humannerf, liu2023robust, song2023nerfplayer} demonstrated the feasibility of learning non-rigid deformations, though they remain computationally demanding and fragile in uncontrolled settings. Gaussian splatting has since been extended to dynamic domains, with method~\cite{luiten2024dynamic, peng2025desire} which introduces time-dependent offsets for
the positions and rotations of Gaussian points.~\cite{yang2024deformable, wu20244d, cao2023hexplane, fridovich2023k, huang2024sc, kwak2025modec, kwon2025efficient} combines a multi-layer perceptron with different sparse spatial deformation methods like planes and control points, boosting the efficiency of training and reducing storage consumption.~\cite{li2024spacetime, park2025splinegs, hu2025learnable, lee2024fully} adopt different forms of polynomial trajectories for deformation, further enhancing the speed compared to the implicit deformation representation.
Graph-based motion modeling approach~\cite{lei2025mosca} employs a set of control node trajectories to handle sparse rigid deformations, providing better representation of complex motion scenarios; we adopt this formulation in our work for its expressiveness.
Recent works have further integrated spatio-temporal modeling into SLAM pipelines, including 4DGS-SLAM~\cite{li20254d}, which jointly estimates poses and separates static from dynamic Gaussian primitives. While these methods improve tracking and rendering quality under motion, they suppress regions affected by blur, occlusion, or exposure instead of modeling them, highlighting the need for uncertainty-aware 4D reconstruction in real-world conditions.


\begin{figure*}[ht] 
    \centering 
    \includegraphics[width=0.98\textwidth]{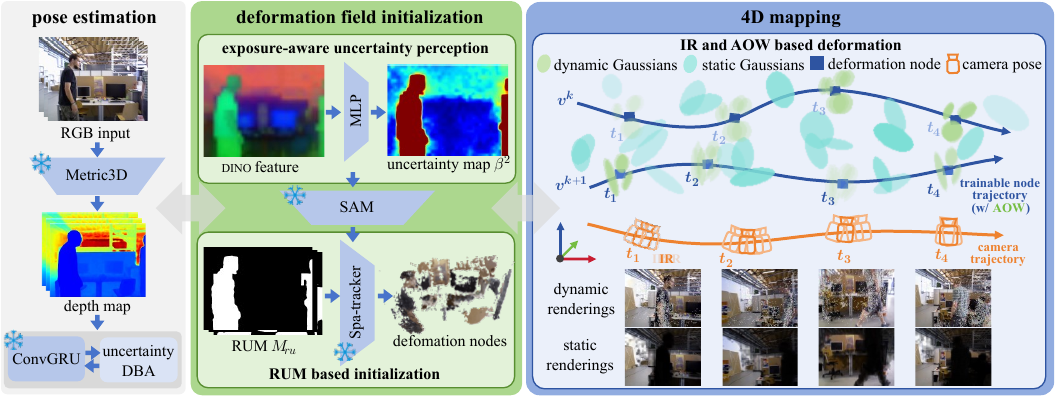}
    \caption{\textbf{Overview of RU4D-SLAM}. 
    RU4D-SLAM operates in three stages: pose estimation, deformation field initialization, and 4D mapping, all of which are closely linked to the uncertainty map $\boldsymbol{\beta}^2$. 
    In the pose estimation stage, $\boldsymbol{\beta}^2$ supports uncertainty-aware DBA tracking. 
    Before 4D mapping, the uncertainty map is combined with SAM to form RUM, within which deformation nodes are initialized as local motion anchors by a pretrained SpaTracker model.
    In 4D mapping, node trajectories are propagated to Gaussians via deformation blending and optimized through IR- and AOW-guided training for joint static and dynamic rendering at each keyframe.
    Snowflake icons denote pre-trained, frozen modules.
    }
    \vspace{-1.3em}

    \label{fig:introduction}
\end{figure*}

\section{Preliminaries} \label{sec:preliminaries}
\paragraph{4D Gaussian Splatting.}
A 4D scene can be rendered using a variant of 3D Gaussian splatting, in which the conventional Gaussians \cite{10.1145/3592433} are extended to evolve over time through locally rigid transformations. Formally, let $\mathcal{G}_s$ be a set of static Gaussians parameterized by their center $\mu_j \in \mathbb{R}^3$, orientation $R_j \in SO(3)$, scale $s_j \in \mathbb{R}^3$, opacity $o_j \in [0,1]$, and color coefficients $c_j$. To capture 4D temporal dynamics, we define the dynamic Gaussians at query frame time $t$ using the motion-scaffold graph\cite{lei2025mosca}, whose nodes {$\mathcal{V} = \{v^{k}\}$} represent 6-DoF trajectories that model the underlying low-rank and smooth motion of the scene:
\begin{equation}
{\mathcal{G}_d(t)} = \{(\mathbb{T}_j(t)\mu_j,\, \mathbb{T}_j(t)R_j,\, s_j,\, o_j,\, c_j)\}_{j=1}^{N},
\label{eq:mosca-dynamic}
\end{equation}
where $\mathbb{T}_j(t)\!\in\!SE(3)$ denotes the pose transformation applied to the $j$-th Gaussian, defined as
\begin{equation}
\mathbb{T}_j(t) =
\mathrm{DQB}\!\left(\{\mathcal{W}(\mu_j,\boldsymbol{t}^i,r^i),\, \Delta \mathbf{Q}^i(t)\}_{i \in \varepsilon(k^*)}\right),
\label{eq:mosca-dynamic-2}
\end{equation}
with 
$\Delta \mathbf{Q}^i(t)=\mathbf{Q}^i(t)[\mathbf{Q}^i(\hat t_j)]^{-1}$ capturing the pose variation of node $v^i$ from its reference time $\hat t_j$ to the current time $t$. Here, $\mathbf{Q}^i(t) = [\mathbf{R}^i(t), \mathbf{t}^i(t)] \in\! SE(3)$ represents the rigid motion of node $v^i$ at frame $t$, and $r^i$ is its spatial control radius. Each Gaussian is bound to the graph node $v^{k^*}$ with the minimum Euclidean distance at the initialization time. $\varepsilon(k^*)$ denotes the edges, that is, the adjacent nodes of node $v^{k^*}$.
The function $\mathrm{DQB}(\cdot)$ performs dual-quaternion blending, interpolating local rigid transformations in a rotation-consistent manner to produce smooth and continuous deformations across time. $\mathcal{W}(\cdot)$  represents the weights of edges. The closer to the adjacent nodes, the greater the weight obtained at the initialization time. Both formulations follow MoSca~\cite{lei2025mosca}.

\paragraph{Uncertainty-Aware Tracking.}

The uncertainty-aware tracking formulation leverages dense bundle adjustment (DBA) formulation~\cite{teed2021droid} with per-pixel uncertainty modeling~\cite{Zheng_2025_CVPR}.
Let the $i$-th keyframe be associated with a camera pose $T_{i}$ and {depth} map $d_i$, and define the edge set $E$ as all keyframe pairs $(i,j)$ with overlapping fields of view connected by optical-flow correspondences. DBA jointly optimizes all camera poses and disparities via
\begin{equation}
\label{eq:dba}
\arg\min_{\{T_{i},\,d_i\}}
\sum_{(i,j)\in E}
\Bigl\|
\hat{p}_{ij} -
f_{ij}\!\bigl(p_i,\,T_{j}^{-1}T_{i},\,d_i\bigr)
\Bigr\|^2_{\hat{\Sigma}_{ij}/\beta_i^2}.
\end{equation}
The warping function $f_{ij}(\cdot)$~\cite{teed2021droid} 
maps pixel $p_i$ from keyframe $i$ to keyframe $j$ using the relative pose $T_{j}^{-1}T_{i}$ 
and {depth} $d_i$. 
$\hat{p}_{ij}$ denotes the {predicted pixel location by pre-trained ConvGRU}, \ie predicted projection of the pixel $p_i$ in the keyframe $j$, 
and $\hat{\Sigma}_{ij}$ represents the flow confidence (covariance) estimated by the flow network. 
The uncertainty map $\boldsymbol{\beta}^2$, whose entries are the per-pixel uncertainties $\beta_i^2$ predicted by a multi-layer perceptron (MLP) based on pre-trained DINO features and jointly optimized within the framework to modulate the residual covariance, 
down-weighting unreliable or dynamic pixels~\cite{Zheng_2025_CVPR}.
 
\section{Methods}
\label{sec:methods}
As illustrated in Figure~\ref{fig:introduction}, RU4D-SLAM centers on embedding uncertainty awareness throughout the entire 4D SLAM pipeline. To achieve this, we introduce three key components—\emph{integrate and render (IR)} (Section \ref{sec:ir}), \emph{reweighted uncertainty mask (RUM)} (Section \ref{sec:rum}), and \emph{adaptive opacity weighting (AOW)} (Section \ref{sec:aow}). Specifically, IR integrates renderings along the camera trajectory, producing a more reliable estimation of $\boldsymbol{\beta}^2$ that directly benefits 4D mapping. Building on this refined uncertainty map, RUM incorporates semantic cues to accurately isolate dynamic regions and guide deformation initialization. AOW then modulates node opacity over time, ensuring coherent dynamic modeling during 4D mapping. Together, these components form a unified uncertainty-aware framework that couples pose estimation, deformation, and 4D reconstruction.

\subsection{Integrate and Render}\label{sec:ir}

Existing methods render $\mathcal{G}_s$ and $\mathcal{G}_d(t)$ at discrete poses, neglecting exposure effects from motion. We propose \emph{integrate and render} that jointly renders $\mathcal{G}(t) = \mathcal{G}_s \cup \mathcal{G}_d(t)$ and accumulates along the camera trajectory over the exposure interval, converting motion-blurred or inconsistently exposed observations into reliable learning signals.

Assuming scene changes have a minor effect during short exposure intervals, we integrate rendered sharp images over the exposure interval $S$ to obtain the blurred image:
\begin{equation}
\boldsymbol{I}(t,T) =
\phi
\int_{\tau \in S}
\mathbb{I}\!\big(\mathcal{G}(t), \Delta T(\tau) T\big)\, d\tau,
\label{eq:iar-cont}
\end{equation}
where $\mathbb{I}(\cdot)$ denotes the instantaneous Gaussian rendering function~\cite{10.1145/3592433}, 
$\phi$ is a normalization factor, and $\Delta T(\tau)$ represents the relative camera motion at sub-time $\tau$.  

In practice, \eqref{eq:iar-cont} is approximated by discretizing the integral:
\begin{equation}
\boldsymbol{I}(t,T) \approx 
\frac{\exp(a)}{|S|} 
\sum_{k=0}^{|S|}
\mathbb{I}\!\big(\mathcal{G}(t),\, \Delta T(k) T\big) + b,
\label{eq:iar-final}
\end{equation}
where the learnable parameters $a$ and $b$ approximate the global exposure coefficient $\phi$, 
with $\exp(a)$ scaling the exposure intensity and $b$ adjusting the ambient bias.  
The relative motion $\Delta T(k) \in \mathrm{SE}(3)$ is interpolated between two trainable control poses~\cite{wang2023badnerf}:
\begin{equation}
\Delta T(k)=\exp\!\left(\frac{k}{|S|}\cdot \log(T_{\rm s}^{-1} T_{\rm e})\right),
\label{eq:dT_k}
\end{equation}
where the start pose $T_{\rm s}$ and end pose $T_{\rm e}$ are initialized from the camera pose $T$ with small Gaussian perturbations and jointly optimized during training. The number of temporal samples $|S|$ is then adaptively adjusted based on the relative motion magnitude between $T_{\rm s}$ and $T_{\rm e}$, controlled by the pre-configured step sizes of rotation and translation.

To optimize 4D mapping, we reduce the photometric discrepancy between $\boldsymbol{I}(t,T)$ and the keyframe RGB observation $\boldsymbol{\tilde I}_t$, where $t \in \{1,\dots,\mathcal{T}\}$ indexes the $\mathcal{T}$ selected keyframes. The Gaussians $\mathcal{G}(t)$ are refined over the $\mathcal{T}$ keyframes using photometric loss:
\begin{equation}
\begin{aligned}
L_{\mathrm{4Dmap}}
&= \frac{1}{\mathcal{T}} \sum_{t=1}^{\mathcal{T}}
\big[
(1 - \lambda_1)\|\mathbf{I}(t,T)-\tilde{\mathbf I}_t\|_1 \\
&\qquad\qquad
+ \lambda_1\,\mathrm{SSIM}(\mathbf{I}(t,T), \tilde{\mathbf I}_t) \\
&\qquad\qquad+ \lambda_2\|\mathbf{D}(t,T)-\tilde{\mathbf D}_t\|_1\big], 
\end{aligned}
\label{eq:lcolor}
\end{equation}
where $\mathbf{D}(t,T)$ is the rendered depth, $\tilde{\mathbf D}_t$ is the depth map of the keyframe predicted by Metric3D \cite{hu2024metric3d}, $\lambda_1$ and $\lambda_2$ balance the photometric, structural and geometric terms, and $\mathrm{SSIM}(\cdot) $computes structural similarity~\cite{Zheng_2025_CVPR}. Keyframes are selected based on optical-flow magnitude and covisibility variation \cite{Zheng_2025_CVPR, matsuki2024gaussian}. A new keyframe is inserted when the flow displacement or Gaussian-overlap change exceeds a threshold, indicating a notable viewpoint shift.

\subsection{Reweighted Uncertainty Mask} \label{sec:rum}
Our \emph{reweighted uncertainty mask} separately handles static and dynamic regions. In \emph{exposure-aware reweighting}, we adapt~\eqref {eq:iar-final} for static rendering, \ie rendering only the static subset $\mathcal{G}_s$ under identical exposure, enabling per-pixel uncertainty to handle inconsistent exposure. In \emph{semantic-guided reweighting}, we introduce uncertainty masks enriched with semantic motion cues to estimate per-pixel uncertainty for dynamic regions.

\paragraph{Exposure-Aware Reweighting.}

Let $\boldsymbol{I}(T)$ be the static rendering and $\boldsymbol{D}(T)$ the corresponding depth map at camera pose $T$. The MLP is jointly trained to predict the uncertainty $\beta_i^2$ of the $i$-th pixel via:
\begin{equation}
L_{\text{u},i} =
\frac{\mathrm{SSIM'}\big(\boldsymbol{I}_i(T), \tilde{\boldsymbol{I}}_i\big)
+ \lambda_{1} \lVert \boldsymbol{D}_i(T) - \tilde{\boldsymbol{D}}_i \rVert_{1}}
{\beta_i^{2}},
\label{eq:uncer_iar}
\end{equation}
where $\tilde{\boldsymbol{I}}_i$ denotes the value of the $i$-th pixel in the RGB input $\boldsymbol{\tilde I}$, and $\tilde{D}_i$ the depth of the $i$-th pixel in the depth map $\boldsymbol{\tilde{D}}$ predicted by Metric3D \cite{hu2024metric3d}. $\mathrm{SSIM'}(\cdot)$ is the modified structural similarity loss~\cite{Zheng_2025_CVPR,ren2024nerf}.

\paragraph{Semantic-Guided Reweighting.} Next, the predicted per-pixel uncertainty $\beta_i^2$ is leveraged to guide semantic segmentation for motion extraction. As the first step, an uncertainty mask $\boldsymbol{M}_u$ is constructed as a binary map, with $M_{\mathrm{u},i}$ denoting its binary value at $i$-th pixel:
\begin{equation}
\boldsymbol{M}_{\mathrm{u},i} = \mathbf{1}(\beta_i^2 > \delta_u),
\end{equation}
where $\mathbf{1}(\cdot)$ is the indicator function and $\delta_u$ is a threshold hyperparameter.
However, $\boldsymbol{M}_u$ may only partially capture motion regions.
To recover complete motion objects, sample points within $\boldsymbol{M}_u$ are used as prompts together with the corresponding RGB input for a pre-trained SAM~\cite{kirillov2023segment} model.
The SAM model produces $K$ segmentation candidates $\{\boldsymbol{M}_{\mathrm{s}}^{k}\}_{k=1}^K$, and
the $k$-th candidate, spatially aligned with $\boldsymbol{M}_u$, is evaluated by its overlap ratio as:
\begin{equation}
\rho^{k} = \frac{\sum_i \boldsymbol{M}_{\mathrm{s},i}^{k} \boldsymbol{M}_{\mathrm{u},i}}{\sum_i \boldsymbol{M}_{\mathrm{s},i}^{k}}.
\end{equation}
Candidates showing strong overlap are merged with $\boldsymbol{M}_u$ to produce the final reweighted uncertainty mask:
\begin{equation}
\boldsymbol{M}_{\mathrm{ru}} = \boldsymbol{M}_u \cup \bigcup_{\{k \,|\, \rho^{k} > \delta_{ru}\}} \boldsymbol{M}_{\mathrm{s}}^{k},
\label{eq:rum}
\end{equation}
where $\delta_{\rm ru}$ is the hyperparamter control merging threshold. 
The resulting $\boldsymbol{M}_{\mathrm{ru}}$ exploits the fact that high-uncertainty regions in static mapping typically indicate motion, allowing these regions to be isolated for 4D mapping.

\subsection{Adaptive Opacity Weighting}\label{sec:aow}

Accurate initialization of deformation nodes is crucial for stable 4D reconstruction. However, this step can be fragile: node initialization relies on 2D pixel trajectories estimated by a pre-trained model such as Spa-tracker~\cite{xiao2024spatialtracker}, which may become unreliable under fast motion, occlusion, or appearance ambiguity. Such trajectory errors can misinitialize deformation nodes, causing Gaussians to be warped to incorrect positions across frames and introducing temporal artifacts. To address this, we refine the MoSca graph with \emph{adaptive opacity weights} that allow each node to adjust its visibility over time.

With $\boldsymbol{M}_{\mathrm{ru}}$, we sample $K$ motion nodes $\{v^{k}\}_{k=1}^{K}$ within the regions identified as dynamic. Each node is extended with a learnable time-varying opacity weight $\hat w_{o}^k(t)$, in addition to the original MoSca attributes:
\begin{equation}
v^{k} = 
\big(\{\mathbf{Q}^{k}(t)\}_{t=1}^{\mathcal{T}},
     \{\hat w_{o}^{k}(t)\}_{t=1}^{\mathcal{T}},
     r^{k}\big).
\end{equation}
The temporal size of each node is determined by the maximum number of keyframes $\mathcal{T}$. For each Gaussian in set $\mathcal{G}_d(t)$, its final rendering opacity is weighted by:
\begin{equation}
\tilde{o}_j(t) = \sigma(w_{o,j}(t))\,o_j,
\label{eq:alpha_traj}
\end{equation}
where $o_j$ is the base opacity in \eqref{eq:mosca-dynamic} and $\sigma(w_{o,j}(t))\!\in(0,1)$ represents the time-dependent visibility confidence, with $\sigma(\cdot)$ denoting the Sigmoid function. The term $w_{o,j}(t)$ aggregates opacity weights from neighboring nodes using the edge weights:
\begin{equation}
w_{o,j}(t)=\sum_{i\in\varepsilon(k^*)}\mathcal{W}(\mu_j,\boldsymbol{t}^k,r^k)\,\hat{w}_{o,j}^{k}(t).
\end{equation}

\section{Experiments}
In this section, we first describe the experimental setup, including three benchmarks covering both controlled and in-the-wild dynamics, along with training and evaluation details. We then perform an ablation study to analyze the contribution of each key component: \emph{integrate and render (IR)} for exposure-aware rendering,\emph{ reweighted uncertainty mask (RUM)} for motion segmentation, and \emph{adaptive opacity weighting (AOW)} for temporally consistent mapping.
Finally, we present RU4D-SLAM’s performance on 4D reconstruction and tracking, compared to recent Gaussian-based SLAM baselines in dynamic scenes.

\subsection{Settings}
\paragraph{Datasets.}
The TUM RGB-D dataset~\cite{sturm12iros} contains dynamic indoor sequences and serves as a standard benchmark for RGB-D SLAM evaluation.
The Bonn RGB-D dataset~\cite{palazzolo2019refusion,bonn_rgbd_dynamic_dataset} consists of 24 dynamic indoor scenes with object motion and frequent occlusions.
The Wild-SLAM dataset~\cite{Zheng_2025_CVPR} includes 7 iPhone-captured sequences (4 outdoor and 3 indoor) recorded real-world human activities.
For simplicity, we refer to these datasets as TUM, Bonn, and Wild-SLAM.
We select 6 sequences from TUM and 9 from Bonn that exhibit noticeable dynamic motion. Following WildGS-SLAM~\cite{Zheng_2025_CVPR}, we use RGB inputs and estimate depth using Metric3D~\cite{hu2024metric3d} across all datasets.

\paragraph{Training and Evaluation.}

Our approach focuses on reweighting uncertainty to make 4D Gaussian splatting more robust to motion, exposure, and dynamic objects. During training, we incorporate existing regularization techniques in addition to the proposed objectives in~\eqref{eq:uncer_iar} and~\eqref{eq:lcolor} to improve the training performance. Specifically, for static mapping, we add regularizations for feature loss~\cite{Zheng_2025_CVPR} alongside~\eqref{eq:uncer_iar} to minimize the variance of predicted uncertainty for those having high similarity, and prevent uncertainty from being infinitely large, using the same hyperparameter setting as~\cite{Zheng_2025_CVPR}. For dynamic 4D mapping, we apply the smoothness regularizer from~\cite{lei2025mosca} to constrain the speed, acceleration of the deformation nodes and the spatial consistency among them, combined with~\eqref{eq:lcolor} and using the same hyperparameter setting as in~\cite{lei2025mosca}. Additionally, we add regularization to node ensure the similarity of AOW at similar times. Unlike~\cite{lei2025mosca}, we remove the consistency check between predicted control node depths and metric depth maps during initialization, as well as the trajectory shaking filter, since these often discard nodes in outdoor scenes or in scenes with large moving objects. Additionally, we set $\delta_{\rm u} = 3.5$ and $\delta_{\rm ru} = 0.2$ to derive RUM.

Evaluation covers both rendering and tracking performance, using the peak signal-to-noise ratio (PSNR), structural similarity index (SSIM), and learned perceptual image patch similarity (LPIPS) for rendering quality, and the absolute trajectory error (ATE) for tracking accuracy. 
All experiments are conducted in PyTorch on a single NVIDIA RTX~4090 GPU.

\subsection{Analysis of IR, RUM, and AOW}
We first conduct ablation studies to assess the qualitative impact of each component, followed by quantitative analysis of their individual contributions.

\begin{figure}[ht]
    \centering
    \includegraphics[width=\columnwidth]{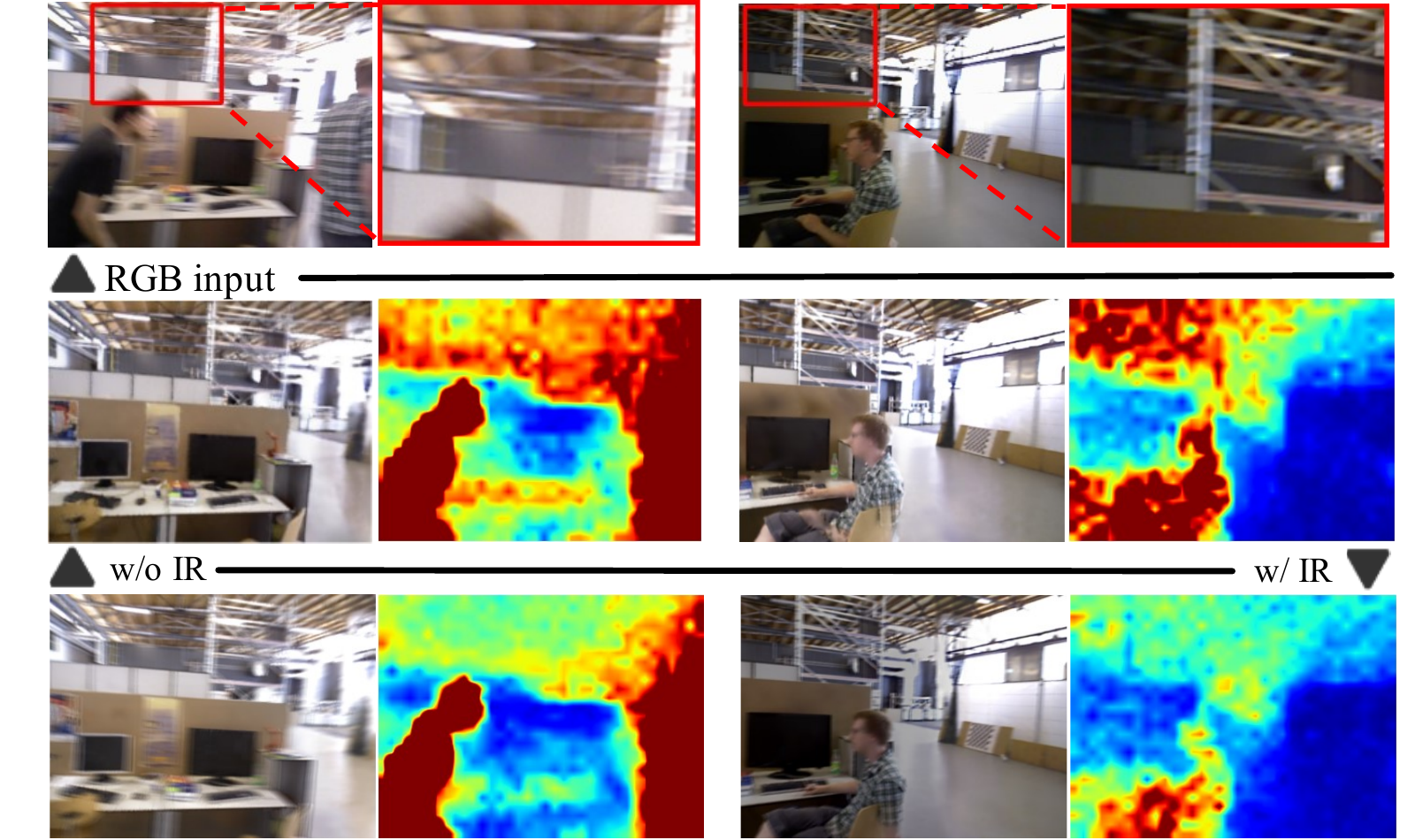}
    \caption{\textbf{Impact of IR on predicted uncertainty map $\boldsymbol{\beta}^2$.} Predicted uncertainty on a sample frame from w\_x and w\_r sequence in TUM dataset. Higher values are shown in red, lower values in blue, with intermediate values transitioning smoothly.}
    \label{fig:exp_uncer_printscreen}
    \vspace{-2em}
\end{figure}

\paragraph{Qualitative Impact.}To evaluate the impact of IR, we visualize the predicted uncertainty map $\boldsymbol{\beta}^2$ from \eqref{eq:uncer_iar} with and without IR. As shown in Figure~\ref{fig:exp_uncer_printscreen}, a frame affected by motion blur causes the MLP to predict high uncertainty in static regions (\cf second row). In contrast, incorporating IR during training produces more stable uncertainty maps with clearer motion–background separation, exhibiting lower predicted uncertainty values in blurred backgrounds. This demonstrates that, by accumulating over the exposure interval, IR stabilizes predicted uncertainty and reduces ambiguity between motion and static regions.

\begin{figure*}[!ht]
    \centering
    \includegraphics[width=\linewidth]{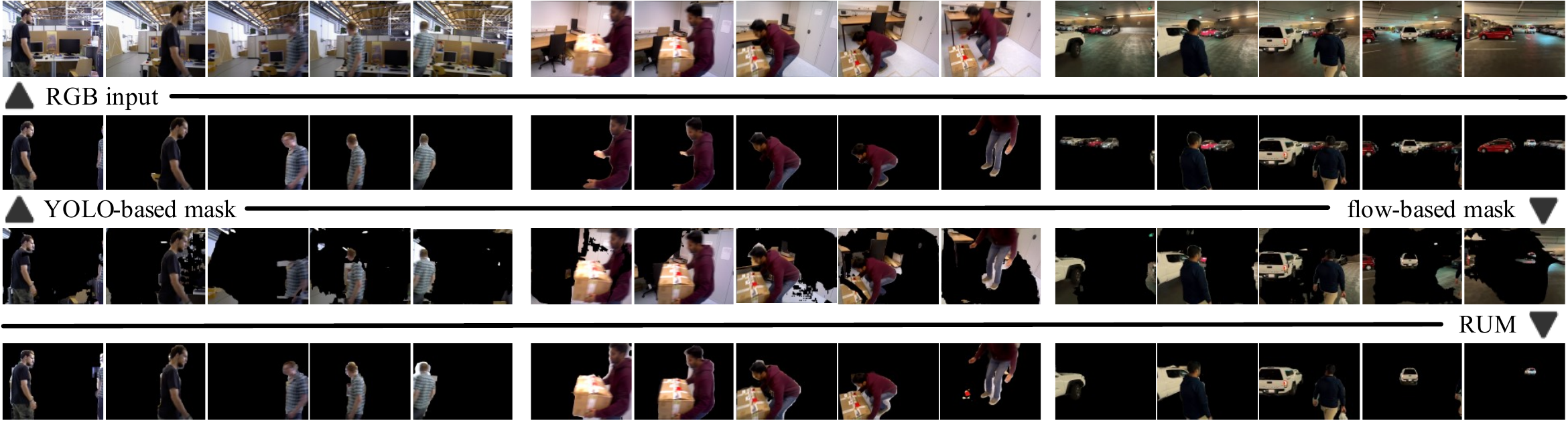}
    \caption{\textbf{Comparison of dynamic object tracking across masks.} Performance of YOLO-based mask~\cite{li20254d}, flow-based mask~\cite{lei2025mosca}, and our RUM on three sample frame sets from w\_x in TUM, pb1 in Bonn and park in Wild-SLAM.
    }
    \label{fig:exp_mask}
    \vspace{-1.5em}
\end{figure*}

The performance of the RUM depends on the quality of the predicted uncertainty map, which benefits from the effectiveness of IR. To demonstrate RUM’s capability in tracking dynamic objects, we compare its performance against YOLO-based masks~\cite{li20254d}, which rely on object priors to localize potential moving instances, and flow-based masks~\cite{lei2025mosca}, which capture pixel-level motion using optical flow.
As shown in Figure~\ref{fig:exp_mask}, RUM consistently focuses on dynamic objects tightly, whereas other masks include background regions. This indicates that RUM suppresses unreliable observations and enhances motion isolation. Guided by predicted uncertainty, RUM preserves complete dynamic objects while effectively excluding background.

\begin{figure}[!ht]
    \centering
    \includegraphics[width=\columnwidth]{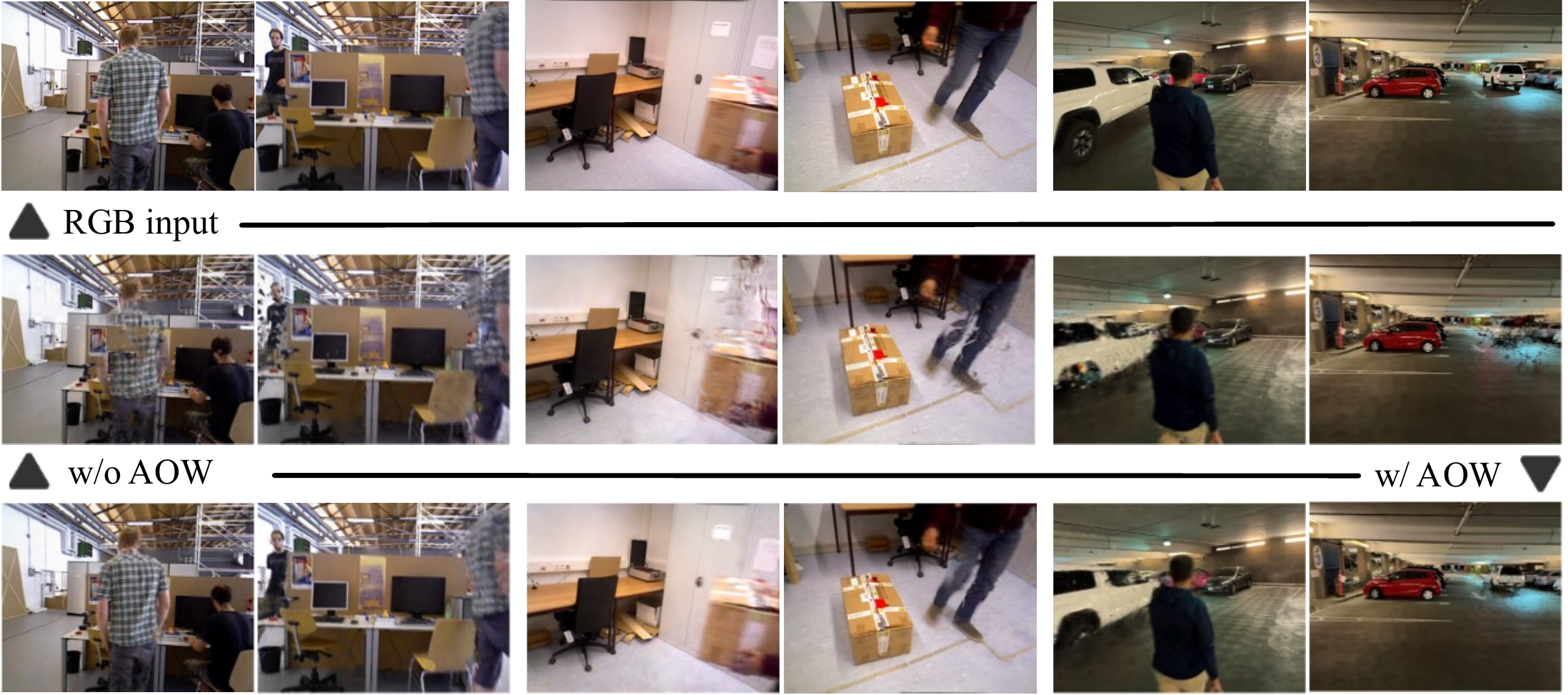}
    \caption{\textbf{Impact of AOW on rendering.}
    Comparison of renderings without and with AOW on frames from w\_x in TUM, pb1 in Bonn and park in Wild-SLAM.}
    \label{fig:ablation_opa}
\end{figure}

To evaluate the impact of AOW, we compare rendering results with and without AOW against the corresponding RGB inputs. As shown in the first column of Figure~\ref{fig:ablation_opa}, without AOW, the dynamic human is not fully rendered, whereas with AOW, it is rendered accurately as in the RGB input. This is because AOW enables motion nodes to modulate Gaussian opacity over time, allowing objects to gradually fade in or out with changing visibility. Consequently, reconstruction errors caused by unreliable initialization of nodes are reduced, and training stability is improved.

\paragraph{Quantitative Analysis.}

We further analyze the contribution of each key component through controlled ablations on TUM. Table~\ref{tab:ablation-tum} reports reconstruction quality (PSNR). Removing AOW or IR clearly degrades performance, with average drops of 0.95~dB and 1.26~dB respectively. AOW prevents Gaussians in motion regions from vanishing and improves temporal consistency (see Figure~\ref{fig:ablation_opa}), while IR enhances robustness to blur and exposure variation.

\begin{table}[!ht]
\centering
\caption{
\textbf{Ablation on TUM (PSNR [dB] $\uparrow$).}
Performance comparison when removing AOW or IR individually. Best per column is \textbf{bold}, second best is \underline{underlined}.}
\label{tab:ablation-tum}
\resizebox{\columnwidth}{!}{
\begin{tabular}{l *{6}{c} c} \toprule
\textbf{Method}& {s\_s} & {s\_x} & {s\_r} & {w\_s} & {w\_x} & {w\_r} & \textbf{Avg.} \\ \midrule
w/o AOW
& 28.70 & \underline{26.18} & 24.73 & 24.62 & \underline{22.66} & \underline{23.11} & \underline{25.00} \\
w/o IR
& \underline{29.11} & 24.29 & \underline{25.17} & \underline{24.88} & 21.78 & 22.91 & 24.69 \\
\textbf{Full model}
& \textbf{29.65} & \textbf{26.34} & \textbf{25.30} & \textbf{26.07} & \textbf{23.94} & \textbf{24.38} & \textbf{25.95} \\ \bottomrule
\end{tabular}
}
\end{table}

We further compare the number of static and dynamic Gaussians generated during training with and without IR. As shown in Table~\ref{tab:ablation-tum-gspoints}, without IR, the system requires more than twice as many Gaussians to represent the uncertain regions caused by motion and exposure blur.

\begin{table}[!ht]
\centering
\small
\caption{Number of dynamic and static Gaussians after training on TUM (in thousands).}
\label{tab:ablation-tum-gspoints}
\resizebox{\columnwidth}{!}{
\begin{tabular}{lcccccccc} \toprule
\textbf{Method} &\textbf{Type} & s\_s & s\_x & s\_r & w\_s & w\_x & w\_r & \textbf{Avg.} \\ \midrule
\multirow{2}{*}{w/o IR} 
& Static  & 378 & 672 & 213 & 347 & 525 & 597 & 459 \\
& Dynamic & 323 & 213 & 514 & 441 & 772 & 893 & 526 \\
\midrule
\multirow{2}{*}{\textbf{Full model}} 
& Static & \textbf{159} & \textbf{274} & \textbf{368} & \textbf{160} & \textbf{263} & \textbf{360} & \textbf{264} \\
& Dynamic & \textbf{100} & \textbf{75} & \textbf{141} & \textbf{112} & \textbf{128} & \textbf{298} & \textbf{142} \\ \bottomrule
\end{tabular}}
\end{table}

\subsection{Rendering and Tracking Results}
\begin{table}[ht]
\centering
\caption{
\textbf{Rendering quality on TUM. (PSNR [dB] $\uparrow$, SSIM$\uparrow$, and LPIPS$\downarrow$)}
$\uparrow$ indicates higher is better; $\downarrow$ indicates lower is better.
Best results are \textbf{bold}, second best are \underline{underlined}.}
\label{tab:psnr-tum}
\centering
\resizebox{\columnwidth}{!}{
\begin{tabular}{llccccccc}\toprule
\textbf{Method} & \textbf{Metric} & \texttt{s\_s} & \texttt{s\_x} & \texttt{s\_r} & \texttt{w\_s} & \texttt{w\_x} & \texttt{w\_r} & \textbf{Avg.} \\ \midrule
\multirow{3}{*}{MonoGS~\cite{matsuki2024gaussian}}
& PSNR$\uparrow$  & 19.95 & 23.92 & 16.99 & 16.47 & 14.02 & 15.12 & 17.74 \\
& SSIM$\uparrow$  & 0.739 & 0.803 & 0.572 & 0.604 & 0.436 & 0.497 & 0.608 \\
& LPIPS$\downarrow$& 0.213 & 0.182 & 0.405 & 0.355 & 0.581 & 0.560 & 0.382 \\
\midrule
\multirow{3}{*}{Gaussian-SLAM~\cite{yugay2023gaussian}}
& PSNR$\uparrow$  & 18.57 & 19.22 & 16.75 & 14.91 & 14.67 & 14.50 & 16.43 \\
& SSIM$\uparrow$  & 0.848 & 0.796 & 0.652 & 0.607 & 0.483 & 0.467 & 0.642 \\
& LPIPS$\downarrow$& 0.291 & 0.326 & 0.521 & 0.489 & 0.626 & 0.630 & 0.480 \\
\midrule
\multirow{3}{*}{SplaTAM~\cite{keetha2024splatam}}
& PSNR$\uparrow$  & 24.12 & 22.07 & 19.97 & 16.70 & 17.03 & 16.54 & 19.40 \\
& SSIM$\uparrow$  & \underline{0.915} & \underline{0.879} & \underline{0.799} & 0.688 & 0.650 & 0.635 & 0.757 \\
& LPIPS$\downarrow$& \underline{0.101} & \underline{0.163} & \underline{0.205} & 0.287 & 0.339 & 0.353 & 0.241 \\
\midrule
\multirow{3}{*}{SC-GS~\cite{huang2024sc}}
& PSNR$\uparrow$  & 27.01 & 21.45 & 18.93 & 20.99 & \underline{19.89} & 16.44 & 20.78 \\
& SSIM$\uparrow$  & 0.900 & 0.686 & 0.529 & 0.762 & 0.590 & 0.475 & 0.657 \\
& LPIPS$\downarrow$& 0.182 & 0.369 & 0.512 & 0.291 & 0.470 & 0.554 & 0.396 \\
\midrule
\multirow{3}{*}{4DGS SLAM~\cite{li20254d}}
& PSNR$\uparrow$  & \underline{27.68} & \underline{24.37} & \underline{20.71} & \underline{22.99} & 19.83 & \underline{19.22} & \underline{22.46} \\
& SSIM$\uparrow$  & 0.892 & 0.822 & 0.746 & \underline{0.820} & \underline{0.730} & \underline{0.708} & \underline{0.786} \\
& LPIPS$\downarrow$& 0.116 & 0.179 & 0.265 & \underline{0.195} & \underline{0.281} & \underline{0.337} & \underline{0.228} \\
\midrule
\rowcolor{cyan!10}
& PSNR$\uparrow$  & \textbf{29.65} & \textbf{26.34} & \textbf{25.30} & \textbf{26.07} & \textbf{23.94} & \textbf{24.38} & \textbf{25.95} \\\rowcolor{cyan!10}
& SSIM$\uparrow$  & \textbf{0.931} & \textbf{0.890} & \textbf{0.844} & \textbf{0.875} & \textbf{0.829} & \textbf{0.813} & \textbf{0.864} \\\rowcolor{cyan!10}
\multirow{-3}{*}{\textbf{Ours}}& LPIPS$\downarrow$& \textbf{0.055} & \textbf{0.090} & \textbf{0.141} & \textbf{0.097} & \textbf{0.143} & \textbf{0.186} & \textbf{0.119} \\ \bottomrule
\end{tabular}}
\end{table}

\begin{table}[!ht]
\setlength{\tabcolsep}{3pt}
\centering
\caption{\textbf{Rendering quality on Bonn. (PSNR [dB] $\uparrow$, SSIM$\uparrow$, and LPIPS$\downarrow$)}
$\uparrow$ indicates higher is better; $\downarrow$ indicates lower is better.
Best results are \textbf{bold}, second best are \underline{underlined}.}
\label{tab:psnr-bonn}
\resizebox{\columnwidth}{!}{
\begin{tabular}{llcccccccccc} \toprule
\textbf{Method} & \textbf{Metric} &
bal1 & bal2 & ps1 & ps2 & sy1 & sy2 & pb1 & pb2 & pb3 & \textbf{Avg.} \\ \midrule
\multirow{3}{*}{MonoGS~\cite{matsuki2024gaussian}}
& PSNR$\uparrow$   & 21.35 & 20.22 & 20.53 & 20.09 & 22.03 & 20.55 & 20.76 & 19.38 & 24.81 & 21.06 \\
& SSIM$\uparrow$   & 0.803 & 0.758 & 0.779 & 0.718 & 0.766 & 0.841 & 0.748 & 0.753 & 0.857 & 0.780 \\
& LPIPS$\downarrow$& 0.316 & 0.354 & 0.408 & 0.426 & 0.328 & 0.521 & 0.428 & 0.372 & 0.243 & 0.342 \\
\midrule
\multirow{3}{*}{\scriptsize Gaussian-SLAM~\cite{yugay2023gaussian}}
& PSNR$\uparrow$   & 20.45 & 18.55 & 19.60 & 19.09 & 21.04 & 21.35 & 19.99 & 20.35 & 21.22 & 20.18 \\
& SSIM$\uparrow$   & 0.792 & 0.718 & 0.744 & 0.719 & 0.784 & 0.837 & 0.750 & 0.768 & 0.814 & 0.769 \\
& LPIPS$\downarrow$& 0.457 & 0.480 & 0.484 & 0.496 & 0.402 & 0.364 & 0.509 & 0.493 & 0.441 & 0.458 \\
\midrule
\multirow{3}{*}{SplaTAM~\cite{keetha2024splatam}}
& PSNR$\uparrow$   & 19.65 & 17.67 & 18.30 & 15.57 & 19.33 & 19.67 & 20.81 & 21.69 & 21.41 & 19.34 \\
& SSIM$\uparrow$   & 0.781 & 0.702 & 0.670 & 0.606 & 0.776 & 0.730 & 0.824 & 0.852 & 0.873 & 0.757 \\
& LPIPS$\downarrow$& \underline{0.211} & 0.290 & \underline{0.283} & 0.331 & \underline{0.227} & 0.258 & \underline{0.191} & \underline{0.165} & \underline{0.152} & \underline{0.233} \\
\midrule
\multirow{3}{*}{\scriptsize 4DGS SLAM~\cite{li20254d}}
& PSNR$\uparrow$   & \underline{25.90} & \underline{22.71} & \underline{21.78} & \underline{20.65} & \underline{23.25} & \underline{25.42} & \underline{23.14} & \underline{24.28} & \underline{25.88} & \underline{23.66} \\
& SSIM$\uparrow$   & \underline{0.874} & \underline{0.838} & \underline{0.832} & \underline{0.820} & \underline{0.812} & \underline{0.892} & \underline{0.845} & \underline{0.873} & \underline{0.886} & \underline{0.852} \\
& LPIPS$\downarrow$& 0.234 & \underline{0.264} & 0.289 & \underline{0.294} & 0.250 & \underline{0.169} & 0.239 & 0.224 & 0.207 & 0.241 \\
\midrule
\rowcolor{cyan!10}
& PSNR$\uparrow$   & \textbf{26.05} & \textbf{24.54} & \textbf{25.66} & \textbf{25.81} & \textbf{27.19} & \textbf{26.84} & \textbf{26.90} & \textbf{26.92} & \textbf{27.10} & \textbf{26.33} \\
\rowcolor{cyan!10}& SSIM$\uparrow$   & \textbf{0.913} & \textbf{0.882} & \textbf{0.898} & \textbf{0.897} & \textbf{0.882} & \textbf{0.924} & \textbf{0.917} & \textbf{0.914} & \textbf{0.919} & \textbf{0.905} \\\rowcolor{cyan!10}
\multirow{-3}{*}{\textbf{Ours}}
& LPIPS$\downarrow$& \textbf{0.133} & \textbf{0.159} & \textbf{0.158} & \textbf{0.143} & \textbf{0.131} & \textbf{0.106} & \textbf{0.129} & \textbf{0.154} & \textbf{0.139} & \textbf{0.139} \\ \bottomrule
\end{tabular}}
\end{table}
\begin{table}[ht]
\setlength{\tabcolsep}{5pt}
\centering
\caption{\textbf{Rendering quality on Wild-SLAM. (PSNR [dB] $\uparrow$, SSIM$\uparrow$, and LPIPS$\downarrow$)}
$\uparrow$ indicates higher is better; $\downarrow$ indicates lower is better.
Best results are \textbf{bold}, second best are \underline{underlined}.}
\label{tab:psnr-wildgs}
\resizebox{\columnwidth}{!}{
\begin{tabular}{llcccccccc}\toprule
\textbf{Method} & \textbf{Metric} & park & piano & shop & street & tow & wall & wand & \textbf{Avg.}  \\\midrule
\multirow{3}{*}{MonoGS~\cite{matsuki2024gaussian}}
& PSNR$\uparrow$  & 21.64 & 14.82 & 14.42 & 17.86 & 16.38 & 14.66 & 13.48 & 16.18 \\
& SSIM$\uparrow$  & 0.777 & 0.353 & 0.326 & 0.601 & 0.667 & 0.361 & 0.302 & 0.484 \\
& LPIPS$\downarrow$& 0.317 & 0.473 & 0.578 & 0.350 & 0.491 & \underline{0.638} & 0.533 & 0.483 \\
\midrule
\multirow{3}{*}{\scriptsize 4DGS SLAM~\cite{li20254d}}
& PSNR$\uparrow$  & \underline{21.96} & \underline{21.13} & \underline{19.57} & \underline{19.63} & \underline{22.62} & \underline{16.55} & \underline{18.62} & \underline{20.01} \\
& SSIM$\uparrow$  & \underline{0.797} & \underline{0.705} & \underline{0.608} & \underline{0.708} & \underline{0.806} & \underline{0.432} & \underline{0.560} & \underline{0.659} \\
& LPIPS$\downarrow$& \underline{0.306} & \underline{0.269} & \underline{0.380} & \underline{0.315} & \underline{0.249} & 0.688 & \underline{0.387} & \underline{0.371} \\
\midrule
\rowcolor{cyan!10}& PSNR$\uparrow$  & \textbf{28.48} & \textbf{24.42} & \textbf{21.64} & \textbf{24.98} & \textbf{27.26} & \textbf{21.69} & \textbf{21.48} & \textbf{24.22} \\
\rowcolor{cyan!10}& SSIM$\uparrow$  & \textbf{0.919} & \textbf{0.833} & \textbf{0.732} & \textbf{0.837} & \textbf{0.909} & \textbf{0.702} & \textbf{0.732} & \textbf{0.809} \\\rowcolor{cyan!10}
\multirow{-3}{*}{\textbf{Ours}}
& LPIPS$\downarrow$& \textbf{0.099} & \textbf{0.101} & \textbf{0.176} & \textbf{0.138} & \textbf{0.089} & \textbf{0.201} & \textbf{0.154} & \textbf{0.137} \\ \bottomrule
\end{tabular}}
\end{table}

\paragraph{Rendering Quality.} 
Tables~\ref{tab:psnr-tum}-\ref{tab:psnr-wildgs} show that RU4D-SLAM achieves consistent state-of-the-art rendering performance across the TUM, Bonn, and Wild-SLAM datasets.
It reaches 25.95\,dB on TUM, 26.33\,dB on Bonn, and 24.22\,dB on Wild-SLAM, with significant improvements in both PSNR and perceptual quality.
These gains align with the results in Figures~\ref{fig:exp_uncer_printscreen}–\ref{fig:ablation_opa}, demonstrating the effectiveness of incorporating our proposed IR, RUM, and AOW into the 4D SLAM pipeline.

\paragraph{Tracking Accuracy.} 
Tables~\ref{tab:ate-tum} and \ref{tab:ate-bonn} present ATE results on TUM and Bonn. RU4D-SLAM achieves the lowest average ATE on both datasets (1.69 cm on TUM and 2.50 cm on Bonn), slightly improving over WildGS-SLAM while remaining competitive with dedicated dynamic-SLAM baselines. These results indicate that incorporating IR for uncertainty refinement leads to a reliable pose estimation under dynamic conditions.

\begin{table}[ht]
\centering
\caption{\textbf{Pose accuracy on TUM (ATE [cm] $\downarrow$).} $\downarrow$ indicates lower is better.
Best values per column are \textbf{bold}, second best are \underline{underlined}. WildGS-SLAM* is re-evaluated without final bundle adjustment to avoid pose errors from dynamic objects.}
\small
\label{tab:ate-tum}
\resizebox{\columnwidth}{!}{
\begin{tabular}{l cccccc c}
\specialrule{1pt}{0pt}{0pt}  
\textbf{Method}& s\_s & s\_x & s\_r & w\_s & w\_x & w\_r & \textbf{Avg.} \\
\specialrule{1pt}{0pt}{0pt}  
MonoGS~\cite{matsuki2024gaussian}  &0.48&1.70&6.10&21.90&30.70&34.20&15.80\\
Gaussian-SLAM~\cite{yugay2023gaussian} &0.72&\underline{1.40}&21.02&91.50&168.10&152.00&72.40\\
SplaTAM~\cite{keetha2024splatam}  &\underline{0.52}&1.50&11.80&83.20&134.20&142.30&62.20\\
RoDyn-SLAM~\cite{jiang2024rodyn} &1.50&5.60&5.70&1.70&8.30&8.10&5.10\\
DG-SLAM~\cite{xu2024dgslam} &0.72 & \textbf{1.00} & 3.49 &0.60 &\textbf{1.60} &4.30 &1.95\\
WildGS-SLAM*~\cite{Zheng_2025_CVPR} & \textbf{0.51}    &1.70     & \textbf{2.19}     &\textbf{0.49}     &2.19  &3.11 & \underline{1.70} \\
4DGS SLAM~\cite{li20254d} & 0.58 & 2.90 & 2.60 & 0.52 & \underline{2.10} & \textbf{2.60} & 1.80 \\
\rowcolor{cyan!10}
\textbf{Ours} & 0.55 & 1.70 & \textbf{2.19} & \textbf{0.49} & 2.16 & \underline{3.06} &\textbf{1.69} \\
\specialrule{1pt}{0pt}{0pt}  
\end{tabular}}
\end{table}

\begin{table}[ht]
\centering
\small
\setlength{\tabcolsep}{1pt}
\caption{\textbf{Pose accuracy on Bonn (ATE [cm] $\downarrow$ ).} $\downarrow$ indicates lower is better.
Best values per column are \textbf{bold}, second best are \underline{underlined}. WildGS-SLAM* is re-evaluated without final bundle adjustment to avoid pose errors from dynamic objects.}
\label{tab:ate-bonn}
\resizebox{\columnwidth}{!}{
\begin{tabular}{l ccccccccc c} \toprule
\textbf{Method}&bal1 & bal2 & ps1 & ps2 & sy1 & sy2 & pb1 & pb2 & pb3 & \textbf{Avg.} \\ \midrule
MonoGS~\cite{matsuki2024gaussian}    &29.60&22.10&54.50&36.90&68.50&0.56&71.50&10.70&3.60&33.10 \\
\scriptsize{Gaussian-SLAM}~\cite{yugay2023gaussian}
&66.90&32.80&107.20&114.40&111.80&164.80&69.90&53.80&37.90&84.30 \\
SplaTAM~\cite{keetha2024splatam}   &32.90&30.40&77.80&116.70&59.50&66.70&91.90&18.50&17.10&56.80 \\
\scriptsize{RoDyn-SLAM}~\cite{jiang2024rodyn}&7.90&11.50&14.50&13.80&1.30&1.40&4.90&6.20&10.20&7.90 \\
DG-SLAM~\cite{xu2024dgslam} &3.70    &4.10     &\underline{4.50}     &6.90     &1.12 &7.71 &24.22    &4.59   &4.67     &6.83\\
\scriptsize{WildGS-SLAM*}~\cite{Zheng_2025_CVPR} & 3.09    &\underline{2.80}  &\textbf{4.35}   &\underline{5.39}     &\underline{0.84}    &\underline{0.55} &\underline{2.07}     &3.16     &2.81     & \underline{2.78} \\
\scriptsize{4DGS SLAM}~\cite{li20254d} & \textbf{2.40} & 3.70 & 8.90 & 9.40 & 2.80 & 0.56 & \textbf{1.80} & \textbf{1.50} & \textbf{2.20} & 3.60 \\
\rowcolor{cyan!10}
\textbf{Ours} & \underline{2.65}    &\textbf{2.45}     &4.78     &\textbf{3.98}     &\textbf{0.78}  &\textbf{0.51} &2.40      &\underline{2.58}    &\underline{2.42}     &\textbf{2.50} \\ \bottomrule
\end{tabular}}
\end{table}

\section{Conclusions}
We introduced RU4D-SLAM, a 4D Gaussian splatting SLAM framework designed to perform robustly on low-quality data affected by motion, exposure, and dynamic objects. RU4D-SLAM enhances 4D Gaussian splatting through \emph{integrate and render (IR)}, \emph{reweighted uncertainty mask (RUM)}, and \emph{adaptive opacity weighting (AOW)}. 
IR accumulates along camera trajectories, handling motion blur and inconsistent exposure, and provides a reliable uncertainty estimate for RUM. Building on this, RUM extends uncertainty-aware tracking to estimate per-pixel reliability in dynamic regions, while AOW uses RUM to initialize deformation nodes and guide learning of time-varying opacity and deformation for temporally coherent reconstruction.

Ablation studies confirm the effectiveness of each component, and RU4D-SLAM outperforms prior Gaussian-based SLAM methods in both rendering quality and tracking accuracy across all benchmarks. 
RU4D-SLAM advances 4D Gaussian splatting SLAM for more complex, real-world scenarios, improving robustness. However, real-world cases can be even more challenging than motion blur, exposure, and dynamic objects. Besides, like all Gaussian splatting–based approaches, to achieve real-time performance remains an important direction for future work.

\section*{Acknowledgements}


This work was supported by the National Key R\&D Program of China (Grant No.~2024YFB4710400). This work also received support from DFG under grant No.~389792660 as part of TRR~248\footnote{CPEC:\url{https://perspicuous-computing.science}}, by DFG grant 547583482, and by Jiangsu Science and Technology Programme BK20251812.

{
    \small
    \bibliographystyle{ieeenat_fullname}
    \bibliography{main}
}


\end{document}